\documentclass[conference]{IEEEtran}
\IEEEoverridecommandlockouts
\usepackage{cite}
\usepackage{amsmath,amssymb,amsfonts}
\usepackage{algorithmic}
\usepackage{graphicx}
\usepackage{textcomp}
\usepackage{xcolor}
\def\BibTeX{{\rm B\kern-.05em{\sc i\kern-.025em b}\kern-.08em
    T\kern-.1667em\lower.7ex\hbox{E}\kern-.125emX}}

\usepackage{eso-pic}
\usepackage{url}
\AddToShipoutPictureBG*{
  \AtPageUpperLeft{%
    \put(0,-40){\raisebox{15pt}{\makebox[\paperwidth]{\begin{minipage}{21cm}\centering
      \textcolor{gray}{This article has been accepted for publication in the proceedings of the \\2022 IEEE International Symposium on Circuits and Systems (ISCAS). DOI: 10.1109/ISCAS48785.2022.9937215} 
    \end{minipage}}}}%
  }
  \AtPageLowerLeft{%
    \raisebox{25pt}{\makebox[\paperwidth]{\begin{minipage}{21cm}\centering
      \textcolor{gray}{ \copyright 2022 IEEE.  Personal use of this material is permitted.  Permission from IEEE must be obtained for all other uses, in any current or future media, including reprinting/republishing this material for advertising or promotional purposes, creating new collective works, for resale or redistribution to servers or lists, or reuse of any copyrighted component of this work in other works.
      }
    \end{minipage}}}%
  }
}
\begin{document}

\title{Parallelizing Optical Flow Estimation on an Ultra-Low Power RISC-V Cluster for Nano-UAV Navigation\\
\thanks{This project has been partially funded by the CHIST-ERA APROVIS3D project through the Swiss National Science Foundation (20CH21\_186991).}
}

\author{\IEEEauthorblockN{Jonas Kühne, Michele Magno, Luca Benini}
\IEEEauthorblockA{\textit{Dept. of Information Technology and Electrical Engineering, ETH Zürich, Zürich, Switzerland} \\
kuehnej@ethz.ch}}

\maketitle

\begin{abstract}
Optical flow estimation is crucial for autonomous navigation and localization of unmanned aerial vehicles (UAV). On micro and nano UAVs, real-time calculation of the optical flow is run on low power and resource-constrained microcontroller units (MCUs). Thus, lightweight algorithms for optical flow have been proposed targeting real-time execution on traditional single-core MCUs. This paper introduces an efficient parallelization strategy for optical flow computation targeting new-generation multicore low power RISC-V based microcontroller units. Our approach enables higher frame rates at lower clock speeds. It has been implemented and evaluated on the eight-core cluster of a commercial octa-core MCU (GAP8) reaching a parallelization speedup factor of 7.21 allowing for a frame rate of 500 frames per second when running on a 50\,MHz clock frequency. The proposed parallel algorithm significantly boosts the camera frame rate on micro unmanned aerial vehicles, which enables higher flight speeds: the maximum flight speed can be doubled, while using less than a third of the clock frequency of previous single-core implementations.
\end{abstract}


\section{Introduction}
Optical flow for velocity and position tracking is increasingly crucial for the navigation of unmanned aerial vehicles (UAV) \cite{Lu2018,wudenka2021}.  Down-facing cameras on UAVs have been used for visual odometry (VO) in combination with other sensors enabling autonomous navigation \cite{Valenti2018}. The optical flow prediction from monocular systems does only contain relative velocity information, therefore such a setup is often combined with an additional distance sensor (e.g. ultrasonic) or acceleration sensor (e.g. IMU) to get an absolute metric velocity estimate \cite{Chen2019}. One of the algorithms that have been proposed is also part of the open-source autopilot framework PX4 under the name PX4FLOW \cite{Honegger2013}. Although PX4FLOW aims to have a high frame rate to enable fast flight of UAVs, today it runs on single-core microcontrollers at 250 frames per second (FPS), only enabling a maximum flight speed of 1.5 meters per second when flying 1 meter above the ground.

\begin{figure}
    \centering
    \includegraphics[width=\linewidth]{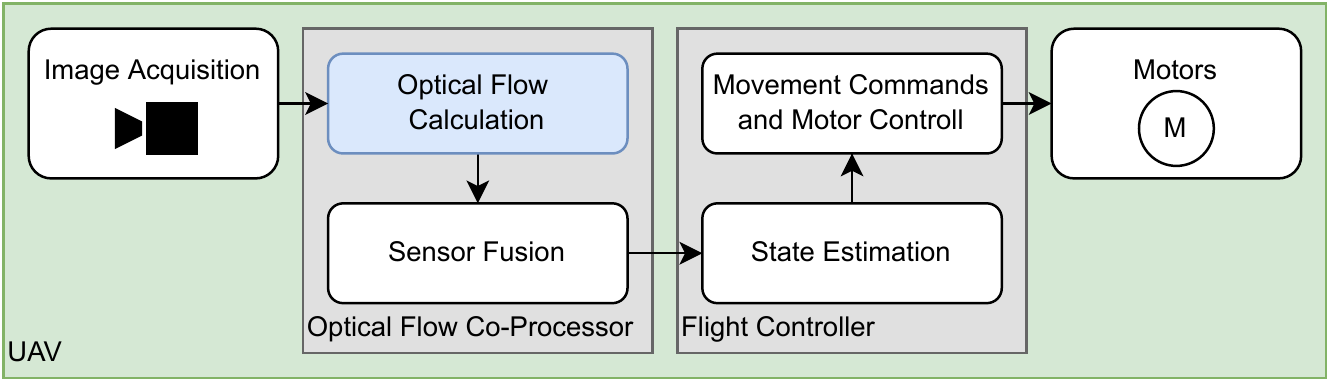}
    \caption{Overview of an optical flow pipeline on UAVs. This paper adresses the algorithm used for the optical flow calculation highlighted in blue.}
    \label{fig:system_overview}
\end{figure}

A typical processing pipeline for optical flow and VO is given in Fig. \ref{fig:system_overview}. The computationally most expensive operation in algorithms like PX4FLOW is the calculation of the actual optical flow, which requires the selection of image features in a first frame and the localization of the matching image features in a second frame. Common optical flow algorithms consider a maximum displacement from one frame to the next one \cite{Kong2021}. The reduced search range limits the computational load on the processor but also restricts the trackable movement speed of the UAV \cite{Honegger2013}. To be able to cope with higher flight speeds, one can either increase the frame rate and keep the search range constant or increase the search range while keeping the frame rate constant. This paper investigates the case of an increased frame rate \cite{Liu2017}.

This paper proposes, develops, and evaluates a parallelized optical flow algorithm on a novel microcontroller of the RISC-V based Parallel Ultra Low Power (PULP) processor family \cite{Pullini2019,Rossi2021}. The specific processor targeted for implementation is the commercially available GAP8 platform by GreenWaves Technology \cite{Flamand2018}. The proposed parallel algorithm is inspired by the PX4FLOW algorithm, which has been proposed and implemented on a single-core ARM Cortex-M4 processor \cite{Honegger2013}. The paper demonstrates and presents how the PX4FLOW algorithm benefits from parallelization and how it can be made more efficient by avoiding recalculations of previously calculated values.

The paper evaluates the proposed algorithm implementing it on the eight cluster cores available on the GAP8 processor. Experimental results demonstrate that the proposed solution yields an excellent multicore speedup of 7.21. The paper analyzes this speedup by considering the theoretical maximum speedup given by Amdahl's law \cite{Hill2008}. Finally, we propose a further optimization to the algorithm that allows it to run another 4.55\,\% faster when using parallelization and 4.94\,\% faster when executed on a single core of the cluster in the GAP8 processor. 

\section{Related Work}
In this section we describe both the related work in the areas of optical flow, especially in the use case of navigating UAVs, as well as the related work in the area of parallel ultra low power microcontrollers, considering both the advances in chip architecture and the applications being implemented on those highly efficient embedded processors.

\begin{figure}[t]
    \centering
    \includegraphics[width=\linewidth]{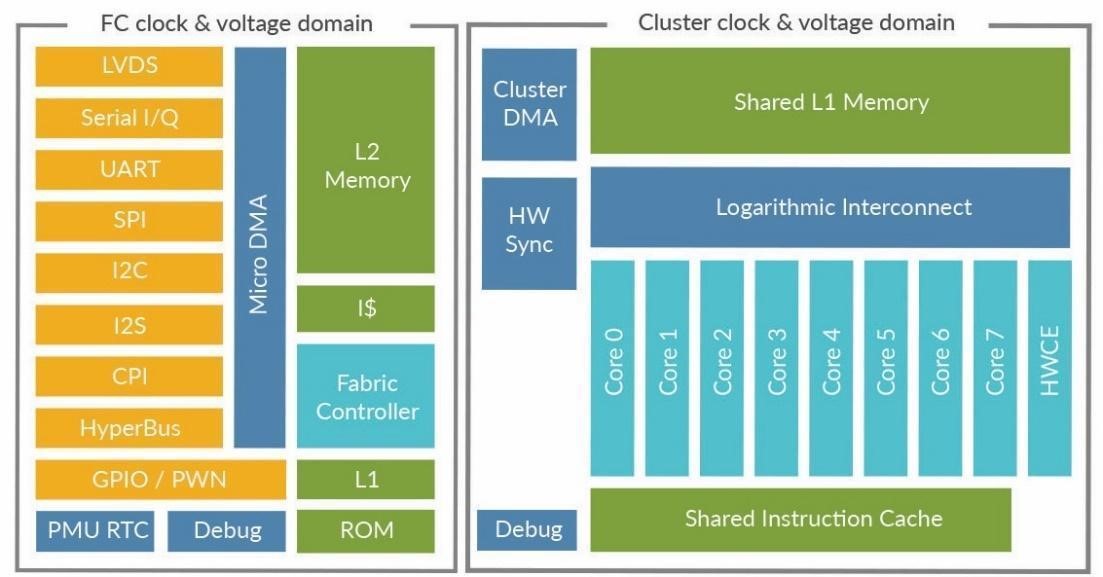}
    \caption{GAP8 Architecture. Source: \cite{Flamand2018}}
    \label{fig:gap8_architecture}
\end{figure}

\subsection{Optical Flow}
Optical flow has been shown to be useful in different tasks on UAVs \cite{Lu2018,wudenka2021,Xiao2021}. Information from optical flow can be used for velocity and position tracking in airborne vehicles, where other odometry information is not available (e.g. in GPS denied areas) \cite{wudenka2021,Valenti2018}. Classical computer vision, using feature detection and matching combined with filtering methods like Extended Kalman Filters, has been used for odometry tasks on UAVs. The used computing platforms range from desktop-grade hardware on the ground \cite{Zingg2010} over specialized onboard FPGA implementations \cite{Watman2011,Liu2017}, to microcontrollers \cite{Honegger2013,Xiao2021}. With the broad adoption of machine learning and deep learning methods, both optical flow predictions \cite{Dosovitskiy2015,Kong2021} as well as end-to-end visual odometry from image frames \cite{Muller2017} have been realized using convolutional neural networks. Additionally, it has been shown that optical flow data can be used for smooth landings of UAVs or collision avoidance in autonomous vehicles, both on drones and cars \cite{Pijnacker2018,Schaub2017}. 

\subsection{Parallel Ultra Low Power Systems}
A focus on edge computing has led to the development of specialized multi-core processors that target Internet of Things (IoT) and edge computing applications with tight power constraints \cite{Pullini2019,Garofalo2021,Rossi2021}.
In recent years the availability of multicore MCUs has favored the research in the area of parallelizable algorithms on resource-constrained embedded hardware \cite{Palossi2019,Djelouat2020}. \cite{Pullini2019} shows how the parallelization of algorithms allows for the use of lower clock rates and lower operating voltages on multicore MCUs. As a linear reduction in voltage leads approximately to a quadratic decrease in the power consumption, parallelized algorithms can potentially lower the energy per instruction, while keeping the throughput of the algorithm. Or put differently to enable higher throughput at the same power level.
Those novel low power processors allow for computationally more intensive tasks such as deep learning on resource-constrained hardware enabling the application on nano drones\,\cite{Mueller2021}.

\section{System Architecture}
The optical flow pipeline for UAVs is illustrated in Fig. \ref{fig:system_overview}. The optical flow co-processor and the flight controller require real-time data processing, which on micro and nano UAVs is done on microcontrollers. As the main contribution of this paper is the proposal and the evaluation of a parallelized optical flow algorithm, the GAP8 application processor by GreenWaves Technologies has been used as a target processor. GAP8 is an ultra low power processor implementing the RISC-V instruction set architecture (ISA), plus additional ISA extensions from the PULP project \cite{Flamand2018}. GAP8 hosts nine compute cores. A single core called fabric controller orchestrates the interaction with the peripherals. The remaining eight cores form the compute cluster that is used for the parallel processing of data. Those cores share a common L1 memory as depicted in Fig.\,\ref{fig:gap8_architecture}.

\begin{figure*}[tb]
    \centering
    \includegraphics[width=0.78\textwidth]{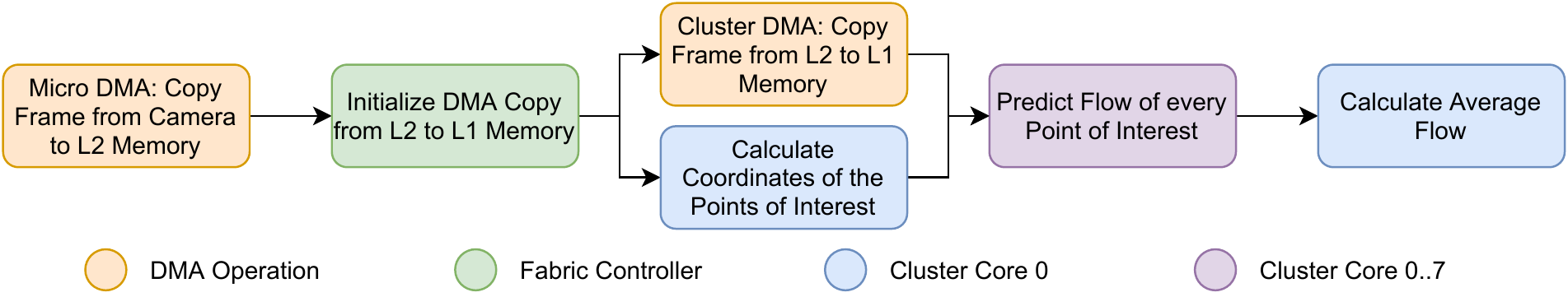}
    \caption{The different processing steps of our implementation. The color-coding indicates which resources are used by which processing step. The DMA copy from L2 to L1 and the calculation of the points of interest is done simultaneously, the flow prediction is only started once both previous tasks have been completed.}
    \label{fig:sequence_diagram}
\end{figure*}

\subsection{Overview}
The processing pipeline on the GAP8 processor is depicted in Fig. \ref{fig:sequence_diagram}. Image frames from a camera get copied into the L2 memory of the GAP8 processor via direct memory access (DMA). Once a new frame is available this new frame gets copied into the L1 memory of the cluster also using DMA. During the copy operation, the coordinates of the points of interest get calculated and the result arrays get initialized. After the initialization which is done on the primary cluster core, the core waits for the completion of the DMA copy operation. Upon completion of the DMA operation, the parallel section of the program is started. Each of the eight cores calculates the displacement of eight different points of interest from frame $n$ to frame $n+1$ for a total of 64 points of interest. The results are then stored and further processed on the primary cluster core. From the 64 flow predictions, separate histograms are built for the displacement in horizontal and vertical directions. The bins with the highest value are then selected as general movement direction, the values are then further optimized by calculating a weighted average from the bins directly adjacent to the selected bin, taking also the evidence of the next two smaller and larger values into account.

\subsection{Proposed Algorithmic Optimizations}
The proposed algorithm is inspired by the PX4FLOW algorithm in \cite{Honegger2013}, which takes 64 predefined, evenly spaced pixel patches from the last image frame $n$ and searches for the corresponding pixel patches in the frame $n+1$. The pixel patches have a size of eight by eight pixels and as a similarity measure the sum of absolute differences (SAD) is used, which is defined in \eqref{eq:sad}. The smaller the SAD the more similar two pixel patches are.

\begin{equation}
    SAD(\mathbf{x_1}, \mathbf{x_2},\Delta \mathbf{x}) = \sum_{\Delta \mathbf{x}} |I_1(\mathbf{x_1} + \Delta \mathbf{x} )-I_2(\mathbf{x_2} + \Delta \mathbf{x})| ,
    \label{eq:sad}
\end{equation}
where $\mathbf{x_1}$ and $\mathbf{x_2}$ are pixel positions in frames 1 and 2 respectively, with $I_1(\cdot)$ and $I_2(\cdot)$ being the grayscale intensity values at the given position for the given frame. $|\cdot|$ denotes the L1 norm and $\Delta \mathbf{x}$ is the patch size over which the SAD calculation shall be performed. With the pixel patches being eight by eight pixels this results in $\Delta \mathbf{x} = (\Delta x,\Delta y)$ with $\Delta x,\Delta y \in [0,7]$. 

\begin{figure}[tb]
    \centering
    \includegraphics[width=\linewidth]{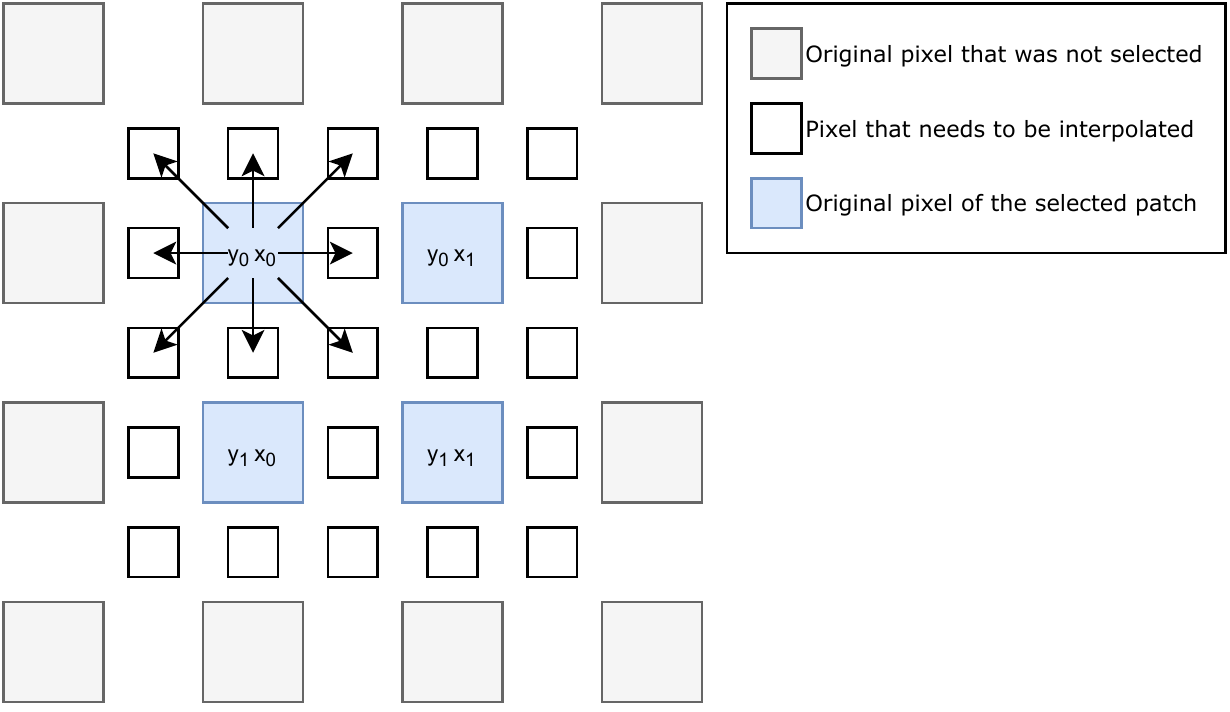}
    \caption{An illustration of the subpixel refinement which is done by bilinearly interpolating half-pixels. For simplicity only a two by two pixel patch is shown instead of an eight by eight patch.}
    \label{fig:pixel_interpolation}
\end{figure}

The search range for the best match in frame $n+1$ is $\pm 4$ in both the x- and y-direction around the pixel coordinate of a point of interest in frame $n$. Once the best match between the frame $n$ and the frame $n+1$ has been determined, PX4FLOW performs a subpixel refinement. In this step, the eight neighboring half-pixel values are calculated using bilinear interpolation as depicted in Fig. \ref{fig:pixel_interpolation}. As the algorithm matches patches of eight by eight pixels between frame $n$ and $n+1$, this requires the calculation of eight refinements per pixel, with 64 pixels.

To reduce the number of calculations, it is important to notice that most of the interpolation values are used multiple times. For example in Fig. \ref{fig:pixel_interpolation} the interpolation value right to the original pixel at $(y_0,x_0)$ is also the left interpolation value of the pixel at $(y_0,x_1)$. Therefore it suffices to calculate these interpolation values once and use them for multiple interpolation directions. Exploiting this feature of the algorithm, the paper proposes two different approaches. 

\begin{enumerate}
    \item The first approach completely avoids recalculations of interpolation values by globally computing all interpolation values for a frame, before the SAD pixel patch matching is performed.
    \item The second approach only avoids local recalculations within the eight by eight pixel patch that is being processed at the time.
\end{enumerate}

\subsection{Parallelization of the Algorithms}
To maximize the utilization of the parallel architecture of GAP8 providing 8 cores, the calculations of the different pixel displacements need to be distributed over the eight cores of the cluster. This is done by distributing the calculations of the 64 different points of interest of the original PX4FLOW algorithm over the eight cluster cores. The calculations per point of interest are not further split. The different points of interest can be treated independently, as no intermediate results need to be known and shared. Once a flow prediction including the subpixel refinement per point of interest is calculated the results are stored in an array.

From the predictions, the average movement direction is calculated involving a floating-point computation. This last step is executed on the primary cluster core.

\section{Experimental Results}
The different approaches have been implemented and evaluated on the GAPuino development board which hosts the GAP8 microcontroller. The GAP8 has an internal performance counter that can be used to measure different metrics (e.g. cycle counter, instruction counter, memory misses, etc.). We were interested in the cycle counter to measure the parallelization speedup as well as the speedup gained from the optimizations in the algorithm. The other metrics gave valuable indications for implementation inefficiencies and were used to find issues when parallelizing the algorithm.

\subsection{Speedup from Optimizations}
Table \ref{tab:subpixel_calculations} shows the number of subpixel refinements for the original algorithm and the two improved versions as well as the resulting number of instructions over the whole algorithm. The implementation where recalculations are avoided locally shows the expected reduction in instructions, while in the implementation that globally calculates the subpixel values only once the number of instructions does not scale as expected. This effect can be explained by the sequence of memory accesses. In the algorithm that optimizes the global calculations, the subpixel refinement is done at the beginning of the parallel section, before any pixel displacements get calculated. Those values are then only used once the optical flow of a certain point of interest gets refined. This requires the subpixel values to get stored in L1 and loaded at a later point. In the locally optimized case, on the other hand, the subpixel values get calculated during the refinement of the optical flow, which allows keeping most of the values in the processor registers and leads to very few values being put in the stack and therefore on L1 memory.

\begin{table}[tb]
    \centering
    \caption{Subpixel Interpolations Required by the Different Algorithms}
    \begin{tabular}{c|c|c|c}
        & \textit{Original} & \textit{Local Optim.} & \textit{Global Optim.} \\
        \hline
        Interpolation Calculations & 49'152 & 19'200 & \textbf{12'288} \\
        Total Instructions & 624.2k & \textbf{585.9k} & 594.3k
    \end{tabular}
    \label{tab:subpixel_calculations}
\end{table}

Both approaches reduced the cycle count of the algorithm. With a single-core speedup of 3.61\,\% (2.73\,\% multicore speedup) for the approach that avoids global recalculations and an even higher single-core speedup of 4.94\,\% (4.55\,\% multicore speedup) when only avoiding recalculations locally. The absolute numbers of clock cycles resulting when executing the instructions from Table\,\ref{tab:subpixel_calculations} are given in Table\,\ref{tab:parallelization_speedup}.

\begin{table}[tb]
    \centering
    \caption{Parallelization Speedup}
    \begin{tabular}{c|c|c|c|c|c|c}
         & \multicolumn{2}{c}{\textit{Original}} & \multicolumn{2}{|c|}{\textit{Local Optim.}} & \multicolumn{2}{c}{\textit{Global Optim.}} \\
        \hline
        Cores & 1 & 8$^{\mathrm{a}}$ & 1 & 8$^{\mathrm{a}}$ & 1 & 8$^{\mathrm{a}}$ \\
        \hline
        Cycles & 717.2k & 99.0k & 681.8k & \textbf{94.5k} & 691.3k & 96.4k \\
        Speedup & - & \textbf{7.24x} & - & 7.21x & - & 7.17x \\
        \multicolumn{7}{l}{}\\
        \multicolumn{7}{l}{$^{\mathrm{a}}$The performance figures are taken from the primary cluster core.}
    \end{tabular}
    \label{tab:parallelization_speedup}
\end{table}

\subsection{Parallelization Speedup}
The parallelization speedup is for all three implementations similar with speedup factors between 7.18 and 7.24 as shown in Table\,\ref{tab:parallelization_speedup}. To put the numbers in perspective the different operations of the algorithm can be separated into a) DMA operation and initialization, b) flow estimation per point of interest (the parallelizable section), and c) histogram and average calculations. The only section that is sped up by the parallelization is b) the flow estimation per point of interest. Therefore we can see a high parallelization speedup in this section as shown in Table \ref{tab:cycle_breakdown} in the row \textit{Flow (8 Cores)}, where the ideal parallelization values are given, with the actual values in brackets, whereas the remaining sections a) and c) are not accelerated. The best implementation regarding cycle count is still the one avoiding local recalculations, although it does not have the highest speedup factor. This can be explained by the fact that the optimization targets the parallelizable section, which therefore becomes faster. The non-parallelizable section remains the same, therefore the speedup factor is lower as in the other implementations although it is the fastest one, this is also reflected in Amdahl's limit as shown in Table\,\ref{tab:cycle_breakdown}.

The parallelized implementations were run on a clock frequency of 50 MHz on the eighth core cluster of the GAP8. With the fastest algorithm, an optical flow prediction can be calculated with up to 529\,FPS. The optical flow predictions can be calculated within 94.5k cycles after a frame has been copied to L2 memory, yielding a latency of 1.89 ms.

\begin{table}[tb]
    \centering
    \caption{Clock Cycle Breakdown}
    \begin{tabular}{c|c|c|c}
        & \textit{Original} & \textit{Local Optim.} & \textit{Global Optim.} \\
        \hline
        Initialization$^{\mathrm{a}}$ & 0.8k & 0.8k & 0.8k \\
        DMA Copy$^{\mathrm{a}}$ & 1.6k & 1.6k & 1.6k \\
        Flow Estimation & 711.5k & \textbf{676.1k} & 685.6k \\
        Averaging & 4.1k & 4.1k & 4.1k  \\
        \hline
        Flow (8 Cores)$^{\mathrm{b}}$ & 88.9k \textit{(93.3k)} & 84.5k \textit{(88.8k)} & 85.7k \textit{(90.7k)} \\
        Speedup Factor$^{\mathrm{c}}$ & 7.578x \textit{(7.24x)} & 7.558x \textit{(7.21x)} & 7.563x \textit{(7.17x)} \\
        \multicolumn{4}{l}{}\\
        \multicolumn{4}{l}{$^{\mathrm{a}}$Both operations \textit{Initialization} and \textit{DMA Copy} are done simultaneously.} \\
        \multicolumn{4}{l}{$^{\mathrm{b}}$Optimal parallelization with the actual cycle counts in brackets.} \\
        \multicolumn{4}{l}{$^{\mathrm{c}}$Amdahl's limit with the actual speedup factors in brackets.}
    \end{tabular}
    \label{tab:cycle_breakdown}
\end{table}

\subsection{Comparison to Related Work}
To have a fair comparison, the original PX4FLOW algorithm from \cite{Honegger2013} has been ported and implemented without modification to GAP8 and is listed as \textit{Original - Single Core} in Table \ref{tab:parallelization_speedup}. It is worth mentioning that the implementation in \cite{Honegger2013} has been evaluated on an STM32F407 microcontroller by ST Microelectronics which uses Cortex-M4F specific single-cycle instructions for the calculations of SAD that do not exist in the ISA of GAP8 and therefore require four cycles on GAP8. As SAD is calculated 91'520 times per frame in 4-byte single instruction multiple data (SIMD) instructions, resulting in 22'880 cycles on Cortex-M4F, the same calculations require 91'520 cycles on a single RISC-V core on GAP8. Therefore it can be estimated, that the original algorithm on Cortex-M4F requires 648'560 cycles per frame, 68'640 cycles (9.57\%) fewer than on a single RISC-V core on GAP8. As the original single-core algorithm on Cortex-M4F runs on a 168\,MHz clock, it can calculate optical flow predictions with just over 250\,FPS as stated in\,\cite{Honegger2013}, which is more than a factor of 2 slower than on GAP8.

The datasheet of the STM32F407 indicates that the current at 168\,MHz is 40\,mA at 1.2\,V resulting in a 48\,mW power draw at full load. In the power analysis of \cite{Palossi2019} the selected GAP8 operating point of this work of 50 MHz for both fabric controller and compute cluster results in a power consumption of 25\,mW. In terms of energy efficiency, the GAP8-based optical flow computation (with 520\,FPS, using 25\,mW at 50\,MHz) is clearly superior to the one on the STM32F407 (with 250\,FPS, using 48\,mW at 168\,MHz) thanks to the combined effect of the more advanced process technology (55\,nm vs 90\,nm) and the parallel execution at a lower frequency.

\section{Conclusion}
The paper presents an approach to speed up the execution of existing algorithms while staying within a tight power envelope by utilizing ultra-low power parallel processors. This paradigm allows bringing higher compute power to embedded devices while adhering to the existing power requirements.

The proposed solutions allow for a faster update rate of the control algorithms by decreasing the latency of optical flow computation by more than a factor of two with respect to the ARM Cortex-M4F clocked at 168\,MHz used in the state of the art PX4 controller.


\bibliographystyle{IEEEtran}
\bibliography{references}

\end{document}